\documentclass[12pt,journal,onecolumn]{IEEEtran}

\ifCLASSINFOpdf
\else
\fi
\usepackage[utf8]{inputenc}
\usepackage{amsmath}
\usepackage{verbatim}
\usepackage{verbatim}
\usepackage{fancyhdr} 
\usepackage{hyperref}       
\usepackage{url}            
\usepackage{booktabs}      
\usepackage{amsfonts} 
\usepackage{nicefrac}       
\usepackage{microtype}      
\usepackage{lipsum}
\usepackage{fancyhdr}       
\usepackage{multirow}
\usepackage{graphicx}
\usepackage{listings}
\usepackage{gensymb}
\usepackage{adjustbox}
\usepackage{setspace}

\begin{document}

\lstset{
	basicstyle=\scriptsize\ttfamily,
	columns=flexible,
	breaklines=true,
	numbers=left,xleftmargin=2em,frame=single,framexleftmargin=1.8em
}
\title{\LARGE \bf
Autonomous Robotic Swarms: A Corroborative Approach for Verification and Validation
}

\author{Dhaminda B. Abeywickrama$^{1,2}$, Suet Lee$^{1}$, Chris Bennett$^{1}$, Razanne Abu-Aisheh$^{1}$, \\Tom Didiot-Cook$^{1}$, Simon Jones$^{1}$, Sabine Hauert$^{1}$, Kerstin Eder$^{1}$%
	\thanks{$^{1}$D. B. Abeywickrama, S. Lee, C. Bennett, R. Abu-Aisheh, T Didiot-Cook, S. Jones, S. Hauert and K. Eder are with the University of Bristol, UK. {\tt\small (firstname.lastname@bristol.ac.uk)} $^{2}$D. B. Abeywickrama is currently with The University of Manchester. *Corresponding author: {\tt\small dhaminda.abeywickrama@manchester.ac.uk}}%
}

\maketitle

\begin{abstract}
The emergent behaviour of autonomous robotic swarms poses a significant challenge to their safety \textit{assurance}. Assurance tasks encompass adherence to standards, certification processes, and the execution of verification and validation (V\&V) methods, such as model checking. In this study, we propose a corroborative approach for formally verifying and validating autonomous robotic swarms, which are defined at the macroscopic formal modelling, low-fidelity simulation, high-fidelity simulation, and real-robot levels. Our formal macroscopic models, used for verification, are characterised by data derived from actual simulations to ensure both accuracy and traceability across different swarm system models. Furthermore, our work combines formal verification with simulations and experimental validation using real robots. In this way, our corroborative approach for V\&V seeks to enhance confidence in the evidence, in contrast to employing these methods separately. We explore our approach through a case study focused on a swarm of robots operating within a public cloakroom.	
\end{abstract}


\section{Introduction} \label{sec:1}
Swarm robotics offers a method for coordinating a large number of robots, inspired by swarm behaviours in nature~\cite{Sahin2005}. 
The collective behaviours of a swarm are not directly engineered into the system. 
Rather, they arise due to interactions among individual robots and their environment, called emergent behaviour~\cite{AERoS}.
Modelling and characterizing emergent behaviour can pose significant challenges~\cite{AbeyCACM2024}, presenting an obstacle to ensuring \textit{assurance} of swarms. 
Assurance tasks encompass adhering to standards, obtaining certification, and conducting verification and validation (V\&V).

\textit{Verification} involves ascertaining that an artifact complies with its specified requirements, while \textit{validation} involves ascertaining that an artifact will effectively perform its intended functions in the real world~\cite{Webster2020}. 
Autonomous robotic systems are inherently complex, thus a range of distinct verification methods, known as heterogeneous verification (e.g., testing, simulation, and formal verification), is required to foster \textit{confidence} in these systems. 
Formal verification encompasses various mathematical techniques (e.g., theorem proving, model checking) used to prove properties concerning a formal representation of a system~\cite{Webster2020}. 
\textit{Model checking} is employed for verifying that a formal model satisfies temporal logic properties by exhaustively exploring the entire state space.  
The outcome of this process usually manifests as a Boolean value, signifying whether the model fulfills a specified property~\cite{Webster2020}. 
When the model fails to satisfy a property, a counterexample (error trace) is generated, describing the  sequence of states that resulted in the violation of the property. 
\textit{Probabilistic model checking} additionally allows the calculation of probabilities that the specified properties will be fulfilled.

\textit{Validation} is a process which, in common with heterogeneous verification, uses a range of techniques (e.g., design reviews, integration testing, and user trials) to establish whether a system is fit for purpose in its intended real-world environment. In this work, validation is performed on our swarm robot platform, DOTS (Distributed Organisation and Transport System)~\cite{Jones2022dots}, via a system integration test using physical hardware in a swarm arena environment representing the intended real-world operational design domain.

Most existing approaches to swarm system modelling focus on simulations, which can be low-fidelity (LF) or high-fidelity (HF). There are several macroscopic formal approaches that aim to analyze overall swarm behaviour~\cite{Liu2010,Konur2012,Lerman2005,Winfield2005,Dixon2011}. However, only a few approaches provide a unified, multi-level representation of swarms (\cite{Endo2023, Egerstedt1999, Martinoli2004, Brambilla2013}). Nevertheless, none of these utilize data obtained from actual simulations to characterize the formal models, nor do they investigate the emergent properties that arise from the inherent qualities of swarms.
\begin{figure}[!t]
	\centering
	\includegraphics[width=0.75\textwidth]{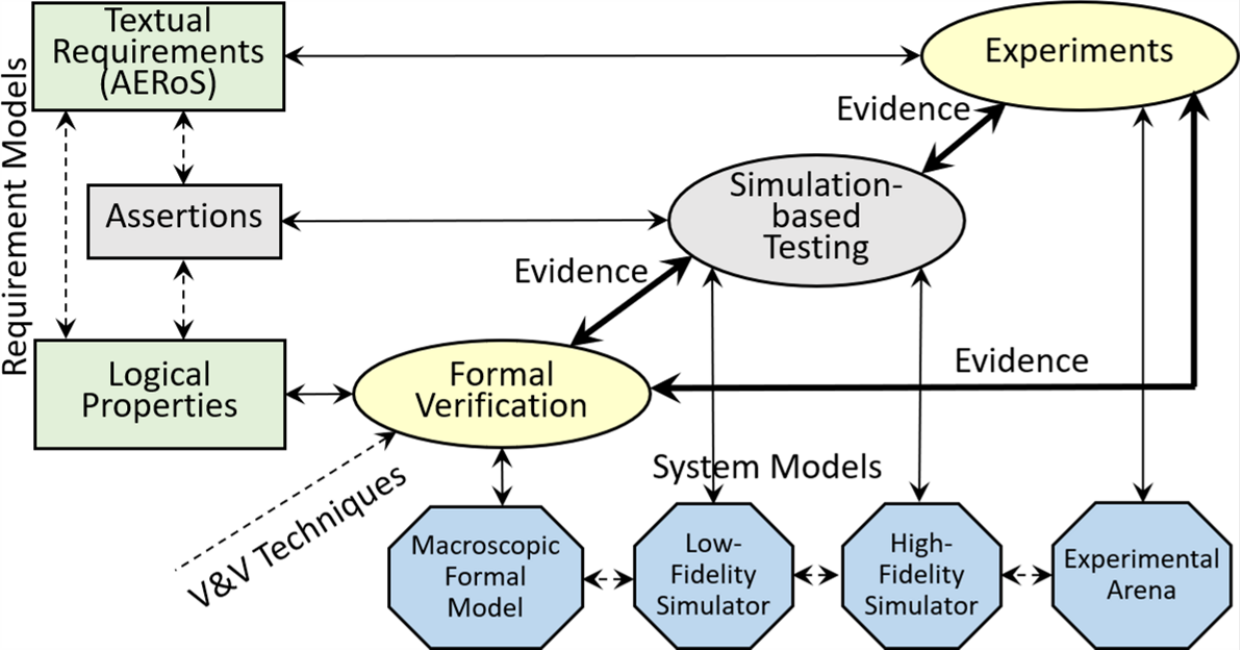}
	\caption{A multi-level corroborative approach for V\&V of autonomous robotic swarms, illustrated by instantiating~\cite{Webster2020} in  this work.} 
	\label{fig:approach}
\end{figure}
The main contributions of this paper are:
\begin{itemize}
\item \textbf{Multi-level Swarm Modelling:} We propose a holistic swarm system modelling approach that spans macroscopic formal modelling, LF simulations, HF simulations, and real-robot levels (see Fig.~\ref{fig:approach}).
\item \textbf{Incorporation of Simulation Data in Formal Models}: Unlike existing approaches, our formal macroscopic models for verification are characterized using data derived from both LF and HF simulations, ensuring accuracy and traceability across different system models.
\item \textbf{Corroborative Approach to V\&V}: We integrate formal verification with experiments involving real robots, aiming to instill greater confidence in the evidence compared to using these techniques in isolation.
\item \textbf{Case Study on a Public Cloakroom}: We apply our approach to a case study involving a swarm of robots operating within a public cloakroom, demonstrating its effectiveness in a practical scenario. 
\end{itemize}      

The paper's structure is as follows. 
Section~\ref{sec:2} provides background information on this work, key related works, and a brief description of the case study. 
In Section~\ref{sec:3} we present our macroscopic formal model.
Section~\ref{sec:4} describes the simulations and the data used to characterize our formal models, while Section~\ref{sec:5} discusses our formal verification approach. 
In Section~\ref{sec:6}, we validate our work using real-world experiments. 
Finally, Section~\ref{sec:7} concludes the paper.

\section{Background}\label{sec:2}
\subsection{Specification: AERoS Process}
The requirements used in this paper are based on the overall specifications derived in~\cite{AERoS}, which proposes a process called AERoS for safety assurance of emergent behaviour in autonomous robotic swarms. AERoS consists of six stages within the emergent behavior lifecycle~\cite{AERoS}. It is derived from the AMLAS process~\cite{Hawkins2021}, which is centered on the assurance of machine learning in autonomous systems. AERoS guidance highlights the critical role of specification in ensuring the safety of the swarm's emergent properties, rather than focusing on the individual behaviors of the robots~\cite{AERoS}.

\subsection{Swarm System Modelling}
Swarm robotic system modelling can occur at four levels of abstraction: macroscopic formal modelling, LF simulations, HF simulations, and real robots~\cite{Endo2023,Baumann2022}. 
The macroscopic level provides a high-level representation of swarm dynamics, particularly beneficial for large-sized swarms. 
Most existing approaches to swarm system modelling focus on simulations.
In LF simulators, details of the environment and real-world physics are abstracted and simplified whilst retaining the most relevant features of the robots in the system~\cite{Lee2022,Lee2023}. In contrast, HF simulators strive to replicate the real-world as closely as possible in order to obtain realistic results to the degree of precision and accuracy required~\cite{Jones2022dots}.

The existing literature in macroscopic modelling can be categorized into three groups: those using rate or differential equations, employing classical control and stability theory for swarm properties, and utilizing other approaches~\cite{Brambilla2013}. 
In \cite{Liu2010}, the authors propose a five-step approach based on differential equations to develop a macroscopic probabilistic model for swarm robotic systems. 
Subsequently, \cite{Konur2012} extend the work in \cite{Liu2010} by using it for verifying the global swarm behaviour with the PRISM model checker~\cite{PRISM_P}. 
Meanwhile, the authors in \cite{Lerman2005} discuss a robot controller designed for a simplified foraging scenario and present a set of differential equations to illustrate the dynamics of their collective behaviour. 
In other approaches, Dixon et al.~\cite{Dixon2011} and Winfield et al.~\cite{Winfield2005} employ linear-time temporal logic to specify properties pertaining to both individual robots and the entire swarm.  
A similar methodology is explored by Konur et al.~\cite{Konur2012} and Brambilla et al.~\cite{Brambilla2013}, wherein they analyze swarm robotic systems using probabilistic model checking. The authors in \cite{Dixon2011} only consider a very small swarm consisting of two or three robots. 

However, only a few approaches offer a unified, multi-level representation of swarms (\cite{Endo2023, Egerstedt1999, Martinoli2004, Brambilla2013}). These follow either a bottom-up or top-down method but do not elaborate on specific V\&V processes. Additionally, they do not utilize data from actual simulations to characterize formal models to ensure accuracy and traceability across different system models, nor do they investigate emergent properties arising from the inherent qualities of swarms.

\subsection{Case Study: The Cloakroom}\label{cloakroom}
Our case study involves a robotic swarm managing a public cloakroom at events attended by 50 to 10,000 people~\cite{Jones2020,AERoS}. 
In this cloakroom, a group of robots assists event attendees in depositing, storing, and retrieving their personal items, such as jackets~\cite{Jones2020}.
Given the swarm's operation in a public environment, the system places a high priority on ensuring public \textit{safety}, which includes measures that prevent harm from collisions and mitigate fire hazards caused by obstructions. 

The scenario presented here is a version of the cloakroom operating in a small window of time (a few minutes), focused at the retrieval of items towards a deposit zone. This scenario was chosen as an example of a logistics task and could be adapted to other scenarios. 
In this scenario, the robots in the swarm initially search for ``carriers", which are platforms used for transporting objects in the cloakroom~\cite{Jones2020}. Robots must position themselves below a carrier to lift it up.
When a robot finds a carrier, it picks it up and transports it to the designated delivery area, employing a random walk while also making efforts to avoid obstacles such as other robots or walls~\cite{Jones2020}. The cloakroom (see Fig.~\ref{fig:LF_sim}) contains three zones with the following usage:

\begin{figure}[!t]
    \begin{center} 
        \adjustbox{trim=0.6cm 0cm 0cm -0.2cm}{
            \includegraphics[width=0.4\columnwidth]{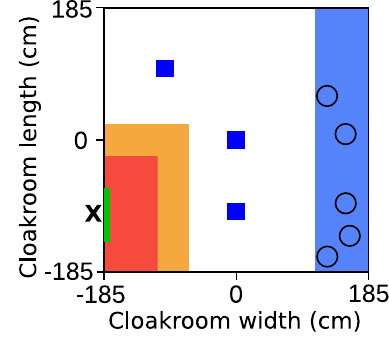}
        }
        \caption{Cloakroom arena: carriers are shown as blue squares, robots as circles. The deposit zone along the right wall is shaded light blue. The fire exit non-blocking zone is shaded red in the bottom left, with the buffer zone shaded amber. The fire exit is marked by an X and shaded green.}
        \label{fig:LF_sim} 
    \end{center}
\end{figure} 
\begin{itemize}
    \item \textbf{Red zone:} To prevent blocking the fire exit, robots must not enter this zone. The dimensions of this zone have been chosen to adequately cover the realistic width of a human doorway (90-100 cm). The red zone is located in the bottom left corner of the cloakroom, with a width of 85 cm and a height of 185 cm. 
    \item \textbf{Amber zone:} This zone is a buffer for the red zone; robots should avoid it or spend minimal time in this zone. The dimensions of this zone have been chosen to allow for a maximum of two robots in this area at any given time. The buffer area is a 50 cm width margin which surrounds the non-blocking area.
    \item \textbf{Deposit area (blue):} Robots deposit carriers in this area. The robots are also initialized within this area at the start of each trial. The deposit area is an 85 cm width zone which runs along the right wall of the cloakroom.
\end{itemize}

\subsubsection{Example Requirements}
Based on the requirements obtained for the cloakroom (refer to Table~1 in~\cite{AERoS}), we identify the following safety-assurance requirements, which are used as examples in this paper. The focus here is on the emergent behaviours of the swarm that could not be predicted from the robot controllers alone.
\begin{itemize}
    \item \textbf{REQ 1: Do not block fire exit}: Several safety assurance levels with increasing criticality are considered, resulting in the following two requirements: (i) The robots in the swarm \textit{shall} not enter the \textbf{red} zone at a fire exit at any time; (ii) Not more than one robot in the swarm \textit{shall} enter the \textbf{amber} zone surrounding a red zone at any time. Additionally, we consider the scenario where a single robot enters the \textbf{amber} zone.
    \item \textbf{REQ 2: Swarm density}: The swarm \textit{shall} ensure fewer than \textbf{10\%} of its robots remain stationary outside the \textbf{delivery site} at any time~\cite{AERoS}. Robots are considered stationary if they have not moved for over \textbf{10 seconds}.
\end{itemize}

\section{Macroscopic Formal Model}\label{sec:3}
In this section, we describe our \textit{macroscopic} model for the overall swarm, adapted from~\cite{Konur2012,Liu2010} for the cloakroom case study. 
At the macroscopic level, the swarm is viewed as a unified entity, offering a high-level representation of swarm dynamics that enhances adaptability, especially in large-sized swarms. 
In our approach, we construct a probabilistic finite state machine (PFSM) for the swarm, which contains the exact same states as those found in an individual robot. 
Additionally, we include a counter to track the number of robots in each state.
Furthermore, we develop a set of `differential equations' for each state within the PFSM of the cloakroom. 
Instead of modelling each robot as a separate PFSM and employing parallel composition to compose the behaviours of all these state machines, we utilize a counting abstraction, also known as a population model. 

\subsection{Probabilistic Finite State Machine Model}
\label{sec:PFSM}
\begin{figure}[!t]
	\centering
	\includegraphics[width=0.75\textwidth]{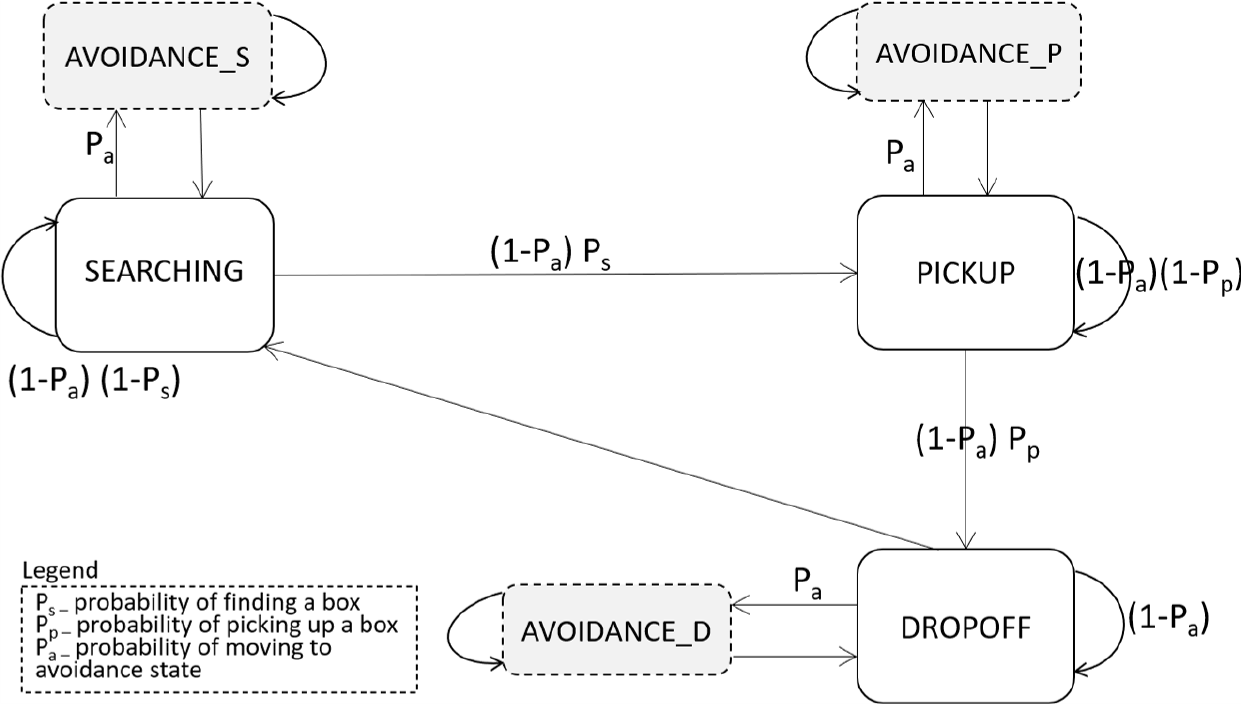}
	\caption{Cloakroom: Probabilistic finite state machine model.} 
	\label{fig:pfsm}
\end{figure}

The PFSM for the swarm consists of six states (refer to Fig.~\ref{fig:pfsm}): SEARCHING, PICKUP, DROPOFF, and their corresponding collision avoidance states—AVOIDANCE\_S, AVOIDANCE\_P, and AVOIDANCE\_D. The associated probability values for these states are as follows: the probability of finding a carrier ($P_s$), the probability of picking up a carrier ($P_p$), and the probability of transitioning to an avoidance state ($P_a$). $T_s$ is the total number of timesteps a robot can be in a particular state.
The robots in the swarm can transition to an avoidance state with a probability of $P_a$ from the SEARCHING, PICKUP, and DROPOFF states. 
Initially, the robots in the swarm are in the SEARCHING state, where they search for carriers by performing a random walk. At each timestep, the robots in the swarm transition to the PICKUP state with a probability of $(1-P_a)P_s$. The robots in the swarm remain in the SEARCHING state with a probability of $(1-P_a)(1-P_s)$. 
From the PICKUP state, robots in the swarm can transition to the DROPOFF state with a probability of $(1-P_a)P_p$, while they stay in the PICKUP state with a probability of $(1-P_a)(1-P_p)$.
If a robot cannot complete a carrier drop-off, it returns to the SEARCHING state or remains in the DROPOFF state with a probability of $(1-P_a)$.

\subsection{Differential Equations}
The differential equations describe the evolution of the average number of robots transitioning between states at a higher swarm level. 
These equations were inspired by the work in \cite{Konur2012,Liu2010}, adapted here for the cloakroom context. 
Currently, timeouts in different states are not considered in this work. These equations have been provided as an example of the type of equations one may define, but these are not intended to be exhaustive.
At time $t$, let:
\begin{itemize}
	\item $N_{S_k}(t)$ be the number of robots in the SEARCHING state for $k$ timesteps $(k \: \epsilon \: {0, ..., {T}_s\!-1})$. 
    Similarly, we can define $N_{P_k}(t)$ and $N_{D_k}(t)$, which are the number of robots in PICKUP and DROPOFF states.
	\item $N_{A_{S_{m,k}}}(t)$ be the number of robots in the AVOIDANCE\_S for $k$ timesteps $(k \: \epsilon \: {0, ..., {T}_s\!-1})$ who have searched for carriers in $m$ timesteps previously $(m \: \epsilon \: {0, ..., {T}_s\!-1})$. Similarly, the number of robots in AVOIDANCE\_P and AVOIDANCE\_D states can be defined.
\end{itemize}

The total number of robots $N$ in the swarm in timestep $k$ can be obtained by equation~\ref{eqn1}: 
\begin{equation} \label{eqn1}
	\begin{aligned}
            N = N_{S_k} + N_{P_k} + N_{D_k}
            + N_{A_{S_{k}}} + N_{A_{P_{k}}} + N_{A_{D_{k}}} 
	\end{aligned}
\end{equation}

Following the PFSM for the cloakroom (Fig.~\ref{fig:pfsm}), several equations can be defined to describe how the average number of robots in each state changes. For example, Equation~\ref{eqn4} determines the number of robots in the SEARCHING state. Similarly, we can define equations for the PICKUP and DROPOFF states. 
\begin{equation}
\begin{aligned}\label{eqn4}
N_{S_k}:N_{S_{0}} (k+1) & = N_{D_{{T}_s-1}}(k) \\
N_{S_{1}} (k+1) & = (1-P_a)(1-P_s)N_{S_{0}}(k)+N_{A_{S_{0,{T}_s-1}}}(k)\\
... \\
N_{S_{{T}_s-1}} (k+1) & = (1-P_a)(1-P_s)
N_{S_{{T}_s-2}}(k) +N_{A_{S_{{{T}_s-2},{T}_s-1}}}(k)\\
\end{aligned}
\end{equation}
Equation~\ref{eqn3} determines the number of robots in the S\_AVOIDANCE state. Similarly, we can define equations for the two other avoidance states. 
\begin{equation} \label{eqn3}
	\begin{aligned}
            N_{A_{S_{m,k}}}: \: \: N_{A_{S_{m,0}}} (k+1) & = P_a N_{S_{m}}(k) \\
		N_{A_{S_{m+1,1}}} (k+1) & = N_{A_{S_{m,0}}}(k)\\ 
		... \\
		N_{A_{S_{m+1,{T}_s-1}}} (k+1) & = N_{A_{S_{m,{T}_s-2}}}(k) 
	\end{aligned}
\end{equation}

\section{Simulations}\label{sec:4}
This section describes our LF and HF simulations developed for the cloakroom case study. First, we present a brief description of the simulation setup.

\subsection{Simulations Setup}
\label{sim_setup}

Carriers have fixed initial locations: (-1, 0), (0, 0), (1, 0), where all coordinates are in meters. Robots are initialized at random at the start of each trial within the blue deposit area. They are modelled to match the properties of the DOTS robots which perform object detection (robot, carrier or wall) via ArUco tags~\cite{Jones2022dots}. Figure~\ref{fig:LF_sim} is a snapshot of the LF simulator and illustrates the scenario. Table~\ref{tab:LF_sim_config} summarizes the configuration.

\subsection{Low-Fidelity Simulation}
The LF simulator, developed in C++, is an abstraction of the cloakroom arena used for real-world trials~\cite{Jones2022dots}. The simulator has been configured to match the real-world setting and DOTS robots as closely as possible in order to minimise the reality gap. Notably, the carriers have been simplified in the LF simulator, and are modelled as objects of circular circumference, 25 cm in diameter.
Robot movement is stochastic and a new heading is selected at random every 0.4 seconds (s). Robots also perform obstacle avoidance upon detection of an object in range of their infra-red laser sensors, and they have an additional 5~cm avoidance margin to prevent physical collisions. Object detection is assumed to have 100\% accuracy and precision in this scenario.
Carriers are removed from the arena as soon as they have been deposited. Figure~\ref{fig:LF_sim} shows a snapshot of the LF simulation environment.
\begin{table}[t!] 
\centering 
    \caption{Configuration of cloakroom and DOTS robots\label{tab:LF_sim_config}} 
    \begin{tabular}{l l l} 
        \hline 
        & Property & Value \\ [0.5ex]  
        \hline 
        Cloakroom & Dimensions & 370 cm x 370 cm \\ [0.5ex]  
        & Number of carriers & 3 \\ 
        & Number of robots & 5 \\ 
        & Carrier diameter & 33 cm \\ 
        \hline

        Robot & Diameter & 25 cm \\
        & Cameras & 4 x 120\degree\hspace{0.1em} FOV video cameras\\
        & & equidistant on perimeter, used\\
        & & for object detection at short and\\
        & &  mid-range, range set to 100 cm\\
        & & in LF;\\
        & & 1 x 120\degree\hspace{0.1em} FOV video camera\\
        & & upward-facing, enables precise\\
        & & positioning under carriers \\
        & Proximity & 16 x IR laser time-of-flight with\\
        & & 3 m range, used for short-range\\
        & & collision avoidance\\
        & Robot max speed & 200 cm/s (real-time) \\ 

        \hline 
    \end{tabular} 
\end{table}

\emph{Data Generation:}
Four datasets were generated for each trial, where at each timestep we collect: (i) the total number of robots in each state where states are defined as in Section~\ref{sec:PFSM}; (ii) the state of each individual robot; (iii) the (x, y) coordinates of each robot (for REQ 1); and (iv) the velocity for each robot and whether a robot is in the delivery zone (for REQ 2). 
A total of 1000 trials were carried out, each lasting 10,000 timesteps (equivalent to a duration of 200 s), which is sufficient to observe the robots retrieving items in our cloakroom.

\subsection{High-Fidelity Simulation}
\begin{figure}[t!]
    \centering
    \includegraphics[width=0.75\columnwidth]{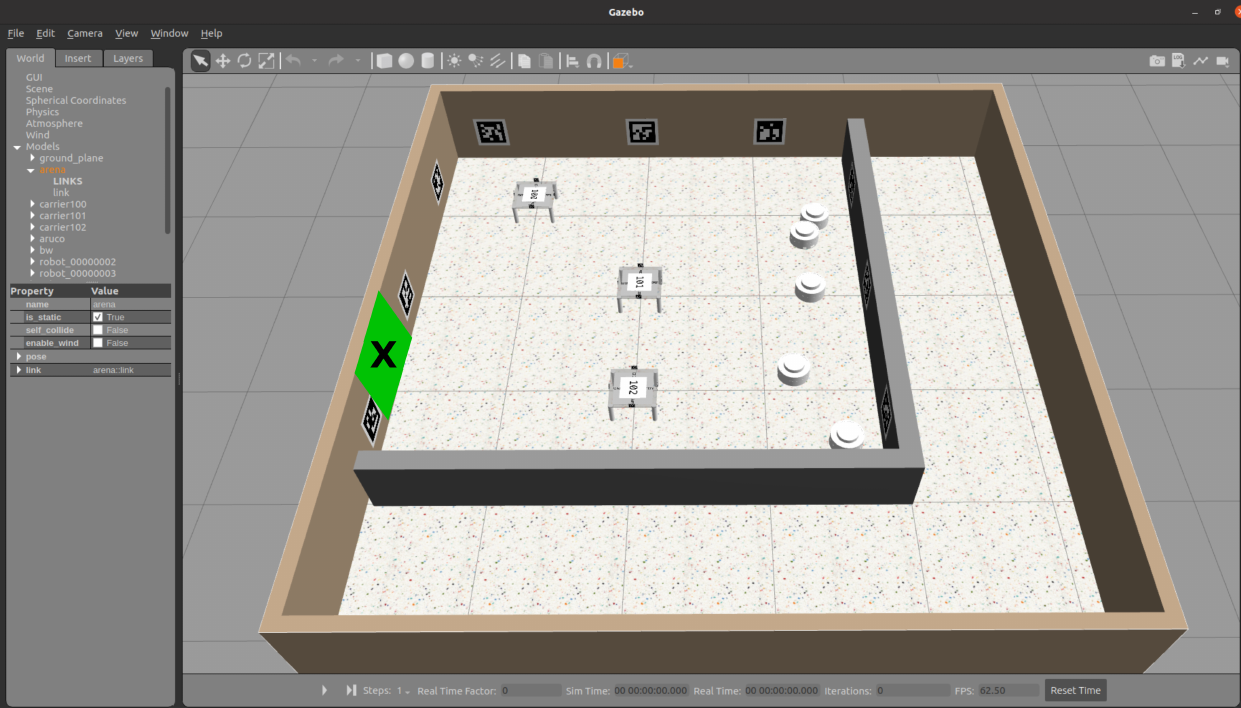}
    \caption{The HF simulator’s graphical interface depicting the initial positions of robots and carriers within the arena boundaries for data collection. The fire exit is shaded green and marked with an X.}
    \label{fig:HF_GUI}
\end{figure}

The HF and LF simulators differ mainly in the detail of the simulation environment. The HF simulator is based on a physics simulator constructed with Gazebo, as shown in Fig.~\ref{fig:HF_GUI}.
This allows modelling the physics of the environment more accurately, enabling realistic interactions between objects, including factors such as gravity, friction, and collision dynamics, which are crucial for robotics and autonomous systems. 
In addition, this approach introduces more realistic noise into the simulation, mimicking the uncertainties and variations present in real-world scenarios, which is essential for robust testing and validation of algorithms in unpredictable environments.
Another difference is that once the robots deposit the carriers in the deposit zone they remain there until the end of the simulation, while in LF they disappear once they have been deposited. 
The code used in the HF simulator is directly transferred to the physical robots, bridging the gap between HF and reality.

A total of 10 HF trials were conducted, each for a duration of 200 s. 
The dataset obtained from these simulations comprises three key pieces of information: the unique identifier for each robot, a timestamp measured in seconds (s), and the corresponding (x, y) coordinates of the robot at that specific moment (for REQ 1).

\section{Formal Verification: Model Checking}\label{sec:5}
In this section, we describe the modelling, specification, and verification of our formal models using the PRISM model checker. A unique feature of our approach is the characterization of formal models using the LF and HF simulation data as introduced in the preceding section.

\subsection{Modelling}
The formal modelling process comprises three steps: data cleaning, data discretization, and modelling in PRISM.

\textit{Data cleaning} involves transforming the raw LF and HF datasets into a form suitable for model checking. For example, (i) in the fire exit requirement, determining whether a robot is in a red or amber zone based on their (x,y) coordinates; and (ii) calculating probabilities for each state by considering the number of robots in each state at each timestep. 
Note, it is assumed that the red zone of the fire exit is enclosed by the amber zone. 
Therefore when checking for violations, a robot in the red zone was counted against the amber and red zone requirements, but a robot in the area adjacent to the red zone was only counted against the amber zone requirement.  

The objective of~\textit{data discretization} is to convert quantitative (continuous) data into qualitative (discrete) forms to manage the number of states in the model. This work employs the Equal Width Discretization (EWD) method, as described in~\cite{Dougherty1995}, to map numerical values into predefined fixed intervals of equal width. 
In the LF simulation, which consists of 10,000 timesteps, we sample data every 50 timesteps to populate our formal models. In the HF simulation, we consider all 200 timesteps of the simulation. 
To illustrate the EWD method, we collect simulation data on the number of robots in each state at each timestep, which allows us to calculate the probability of a robot being in each state at that timestep. The probabilities are divided into five intervals with equal widths (L1 to L5) where L1 denotes the lowest probability, while L5 represents the highest. 
Furthermore, as part of discretization, we also average the data for all the trials, i.e., averaging over 1000 and 10 trials of LF and HF simulation data, respectively. 

At the macroscopic level, the swarm is modelled using continuous-time Markov chains (CTMC)~\cite{PRISM_P}, which capture the evolution of the swarm in each timestep.
This can include information, such as the probabilities of robots in each state; the level of rates (L1 to L5); whether a robot has entered the red zone area of a fire exit; whether more than one robot or a single robot has entered the amber zone of a fire exit; and timestep information. 

\subsection{Specification and Verification}\label{sec:5.2}
In this study, we utilize PRISM's property specification language, specifically employing Continuous Stochastic Logic (CSL)~\cite{PRISM_P}. 
We employ PRISM in verification mode to evaluate all possible runs against CSL formulae. 
A range of properties are created (about 55), for example, to reason about the probability of an event's occurrence (P operator); path properties (e.g., F for eventual or future, G for always or globally); non-probabilistic properties; and reward-based properties (R operator).
In addition, we use the experiments feature of PRISM to graphically illustrate results of automating multiple instances of model checking. 

\subsubsection{Analyzing LF and HF fire exit simulation data}
\begin{figure}[h!]
    \centering 
    \includegraphics[width=0.85\columnwidth]{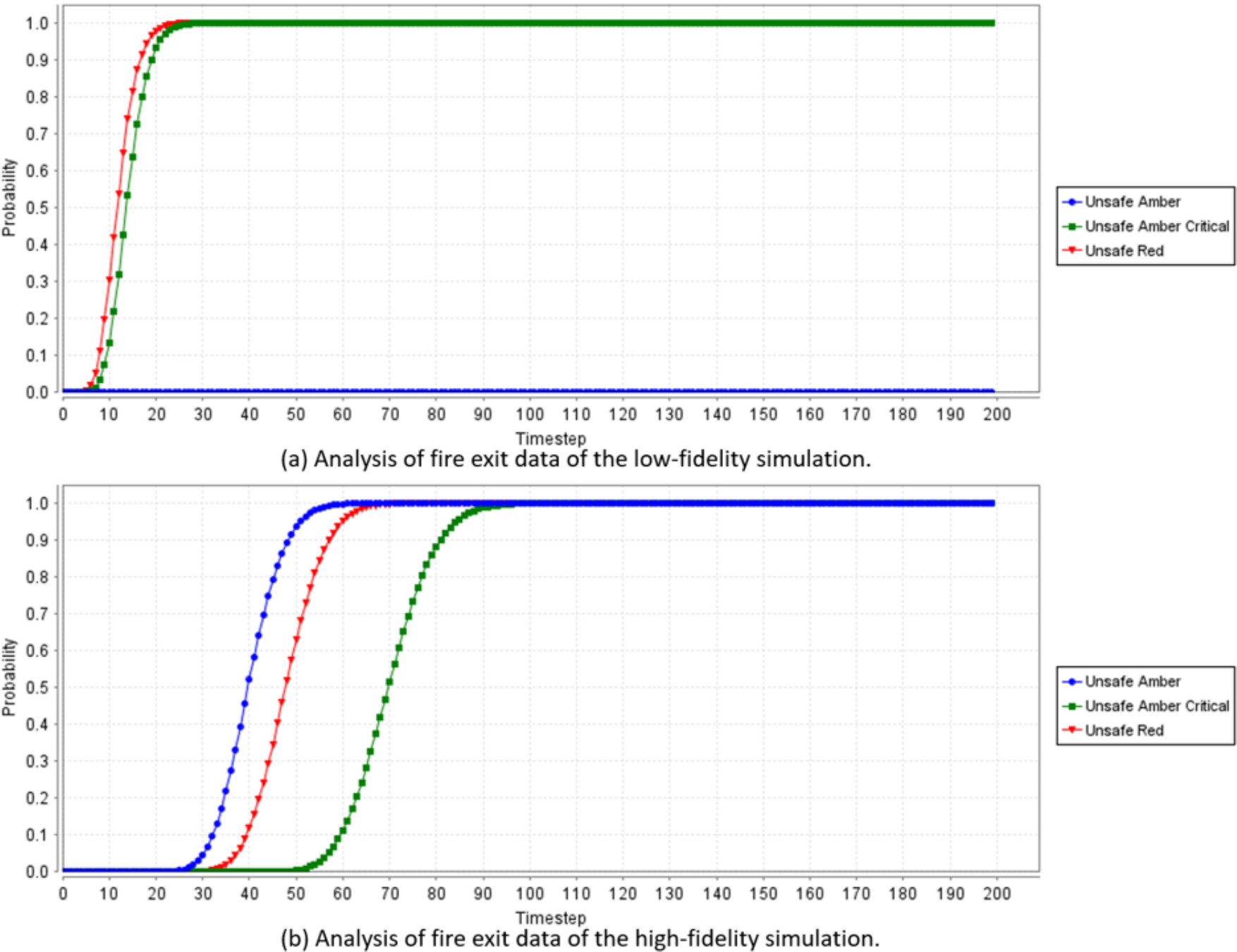}
    \caption{Analysis of LF and HF simulations which show REQ 1 is violated.}
    \label{fig:FER}
\end{figure}

We define several properties (see Listing~\ref{l1}) to determine the probabilities of a robot eventually entering the red zone of the fire exit, and at least two robots (termed `amber critical') or a single robot entering the amber zone. 
For this we use the `F' operator, signifying a property that eventually becomes true at some point along the path.  
We analyze models created using both LF and HF simulation data (see results in Fig.~\ref{fig:FER}). 
Additionally, we employ the `filters' feature of PRISM to return values for all states of the model.
For instance, we print and count states that satisfy the properties where a robot is in the red zone or where more than one robot or a single robot is in the amber zone. Additionally, we calculate the sum and average values of the property over the satisfying states. 
The count, sum, and average values for property checking a robot in the red zone are 14, 14, and 0.07, respectively. In the HF case, these values correspond to 46, 46, and 0.23.

\begin{lstlisting}[frame=single, caption=Key properties to verify a robot entering red and amber zones of the fire exit., captionpos=b, label=l1]
// Probability of a robot entering the red zone within T timesteps
P=? [ F<=T "unsafe_fireexitsblocked" ]
// Probability of at least 2 robots entering the amber zone within T timesteps
P=? [ F<=T "unsafe_amber_critical" ]
// Probability of a robot entering the amber zone within T timesteps
P=? [ F<=T "unsafe_amber" ]
// Sums the value of prop for satisfying states.
filter(sum, P=? [ X "unsafe_red" ])
// The average value of prop over satisfying states.
filter(avg, P=? [ X "unsafe_red" ])
\end{lstlisting}

\subsubsection{Analyzing the progress of the main states against their avoidance states}
\begin{figure}[h!]
    \centering
    \includegraphics[width=0.85\columnwidth]{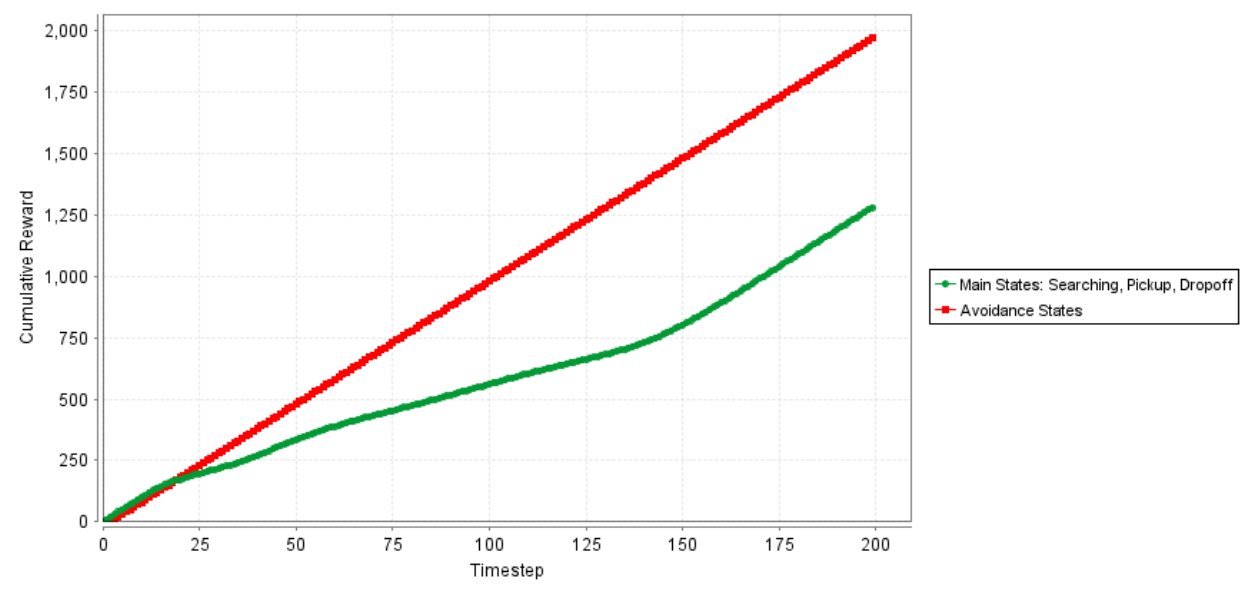}
    \caption{Cumulative rewards for the main states, and their avoidance states. Avoidance state occurs more frequently than a main state in the LF simulation.}
    \label{fig:R_States}
\end{figure}

We also utilize reachability reward properties, specifically cumulative reward properties that associate a reward with each path of a model, constrained by a specific time limit (see Listing~\ref{l2}). These properties have been created to verify if the probability levels of the primary states (e.g., SEARCHING, PICKUP, and DROPOFF) exceed those of their corresponding avoidance states. For this, our model assigns a reward upon the execution of a main and avoidance states, leading to an accumulation of rewards until the simulation's conclusion. It was observed that an avoidance state occurs more frequently than a main state in the LF simulation (see gap between corresponding graphs in Fig.~\ref{fig:R_States}). In addition, several other properties related to instantaneous reward and steady-state reward were formulated.
\begin{lstlisting}[frame=single, caption=Reward properties created to check the cumulative rewards in the main and avoidance states., captionpos=b, label=l2]
R{"main_states"}=? [C<=T]
R{"avoidance_states"}=? [C<=T]
\end{lstlisting}

\subsubsection{Analyzing the rates of robots}
\begin{figure}[h!]
    \centering
    \includegraphics[width=0.85\columnwidth]{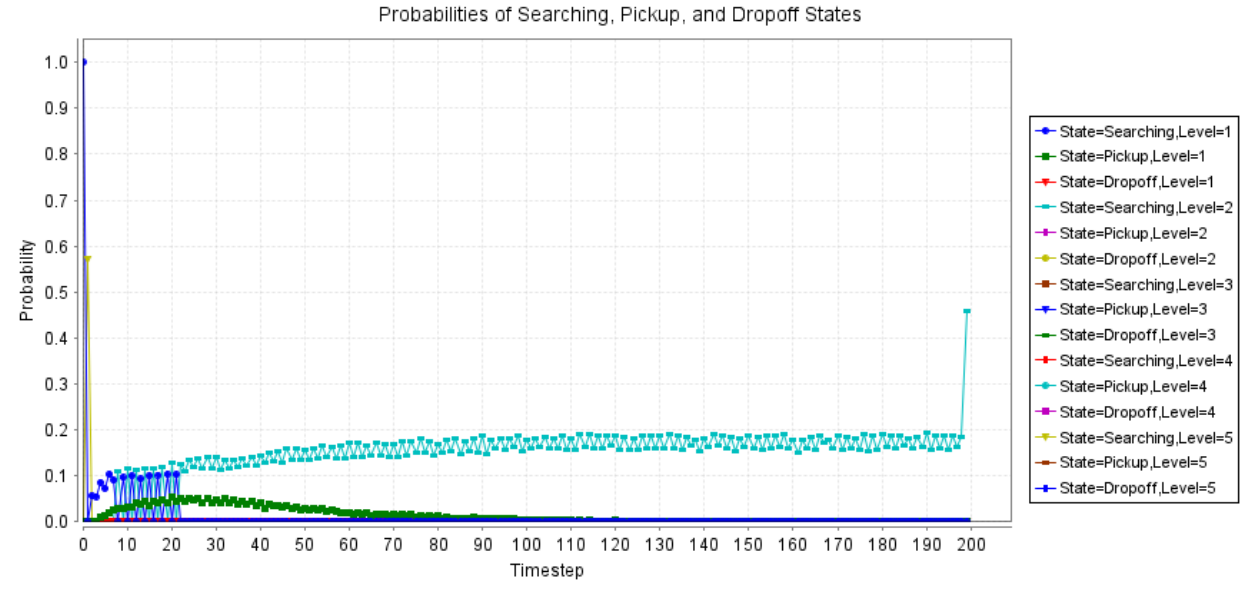}
    \caption{Probabilities of SEARCHING, PICKUP, and DROPOFF states.}
    \label{fig:P_MainStates}
\end{figure}

\begin{figure}[h!]
    \centering
    \includegraphics[width=0.85\columnwidth]{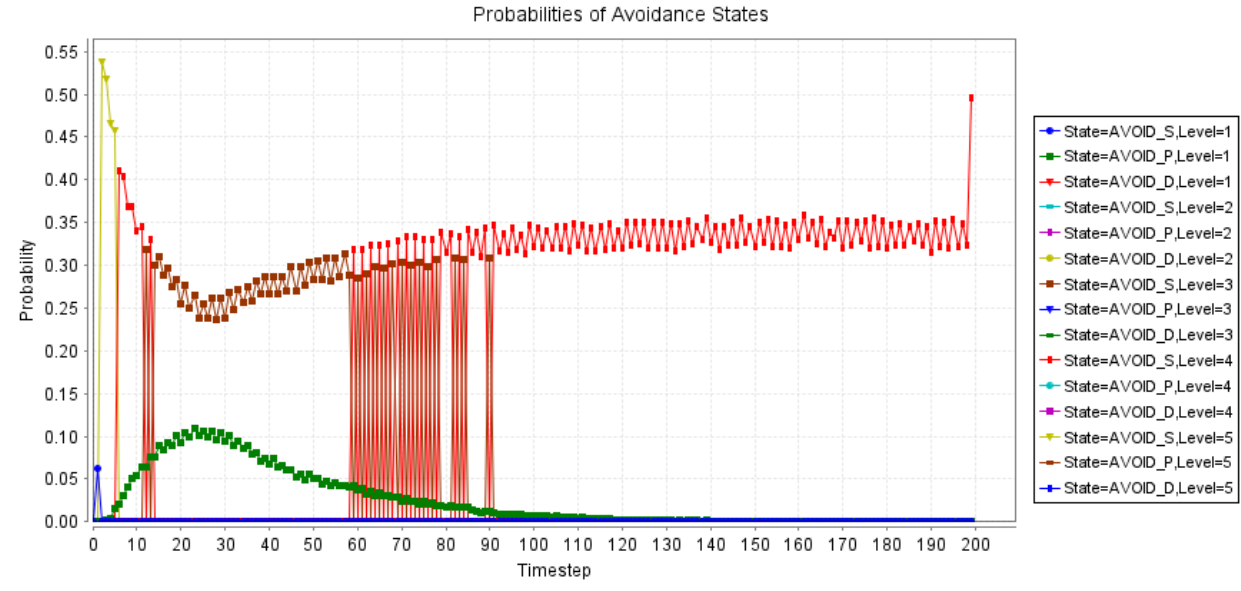}
    \caption{Probabilities of avoidance states are seen to be relatively higher compared to their corresponding main states shown in Fig. 7.}
    \label{fig:A_MainStates}
\end{figure}

\begin{figure}[h!]
    \centering
    \includegraphics[width=0.85\columnwidth]{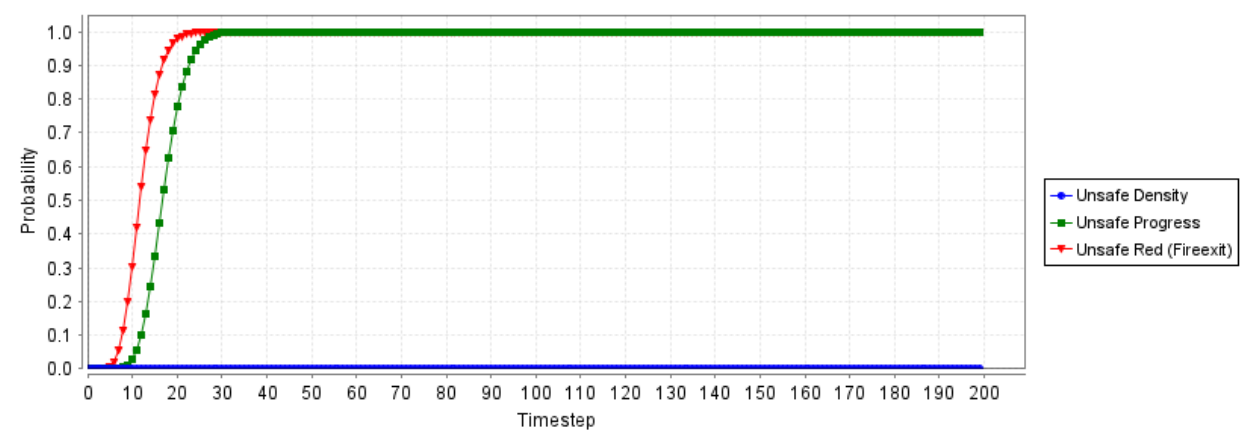}
    \caption{Probabilities of unsafe states (violation of density requirement REQ 2, a robot entering the red zone of the fire exit, and the progression of avoidance states over their main states).}
    \label{fig:UnsafeS}
\end{figure}

Using the experiments feature of PRISM we show the probabilities of the SEARCHING, PICKUP and DROPOFF states (see Fig.~\ref{fig:P_MainStates}), and their avoidance states (see Fig.~\ref{fig:A_MainStates}). Figures~\ref{fig:P_MainStates} and~\ref{fig:A_MainStates} also confirm that the probabilities of the main states are comparatively lower than those of their avoidance states. 
Additionally, we create properties using the bounded variant of path property (U) and F operator to check whether the probability level of the SEARCHING state needs to be at a certain level (e.g., $>=3$) at each timestep in the first half of the simulation, while in the latter half, the DROPOFF state needs to be at a certain level (e.g., $>= 3$) (see Listing~\ref{l3}).  
In addition, we analyze the probabilities of unsafe situations eventually occurring. For example, in the LF simulation: (i) 
 the violation of REQ 2 (density requirement); (ii) a robot entering the red zone of the fire exit; and (iii) the progression of avoidance states over their main states (Fig.~\ref{fig:UnsafeS}).  
 \begin{lstlisting}[frame=single, caption=Properties created to analyze rates of robots., captionpos=b, label=l3]
P=? [ F<=T (s=state&l=level&timestep=T) ]
// Experiment to analyze different states and their eventual probability levels within T timesteps.
P=? [ F[0,99] (s=1&l>=3) ]
// Probability of DROPOFF state being greater than or equal to level 3 between 100 and 199 timesteps.
P=? [ F[100,199] (s=4&l>=3) ]
// Probability of being in SEARCHING state within 99 timesteps (and remaining in DROPOFF state at all preceding time-points) is at least 0.25. 
P>=0.25 [ s=4 U<=99.0 s=1 ]
\end{lstlisting}
\begin{figure} [h!]
	\centering 
        \includegraphics[width=0.85\textwidth]{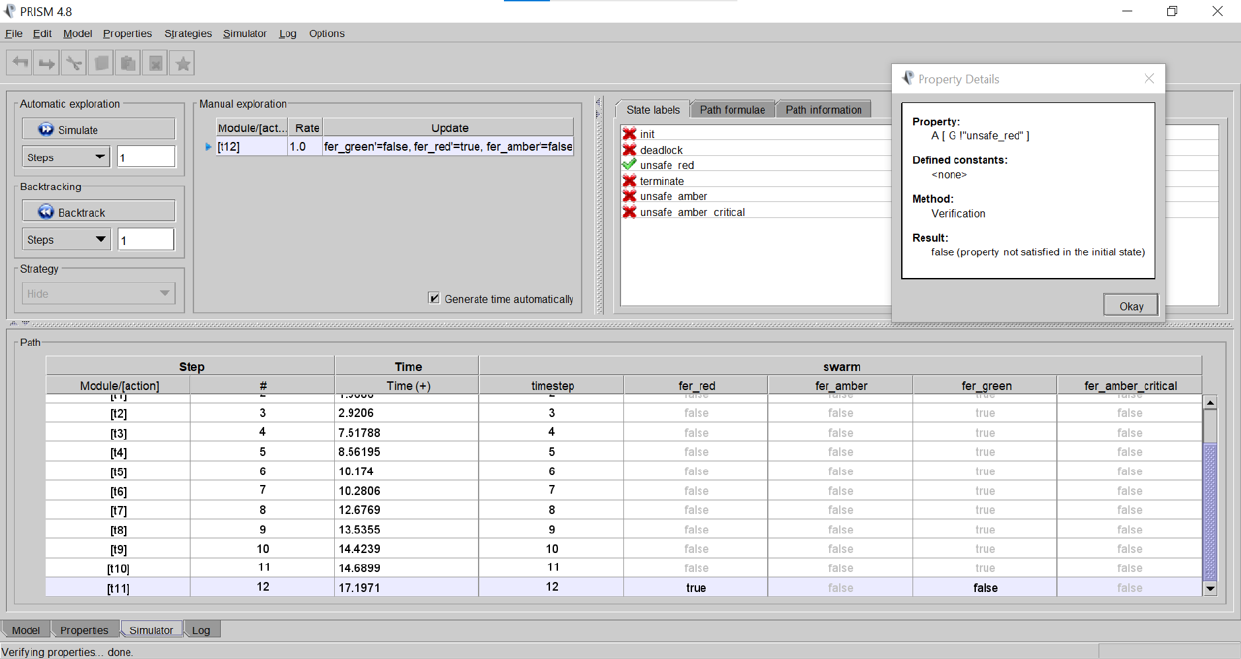}
	\caption{Counterexample trace for property $A [ G !``unsafe\_red" ]$.} 
	\label{fig:fer_2D_p1-2} 
\end{figure}

\subsubsection{Counterexample traces and witness executions: }
We utilize \textit{counterexamples} and \textit{witnesses} generated by CTL properties in the forms of A[ G``inv''] and E [F ``goal''], respectively. In the former, PRISM generates a counterexample illustrating a path that reaches a state where ``inv'' is not true. In the latter, it generates a path reaching a ``goal'' state. 
For instance, let's consider the property A[G $!$``unsafe\_red''] where ``unsafe\_red'' is a label defined on a set of states referred to in the property indicating when a robot enters a red zone of the fire exit (atomic proposition). This property verifies whether the fire exit requirement is met across all states along the path. However, as depicted in Fig.~\ref{fig:fer_2D_p1-2}, this property is violated at timestep 11 in the LF simulation, evident from the respective counterexample trace.

\section{Validation: Experiments with Physical Robots}\label{sec:6}
A system is required to be demonstrably fit for its intended purpose~\cite{ieee2012}. Validation is a process that collects the evidence needed to make this assessment, ideally producing a binary outcome.    
However, ``fit for purpose'' can be subjective, and the evidence used for assessment can be biased, for instance, by the set of tests performed and the test conditions.  Therefore, rather than producing a binary assessment, validation can be described as a process to assess the level of \emph{confidence} that a system meets the needs of its stakeholders.  

\begin{figure}[!t]
	\centering 
	\includegraphics[width=0.75\textwidth]{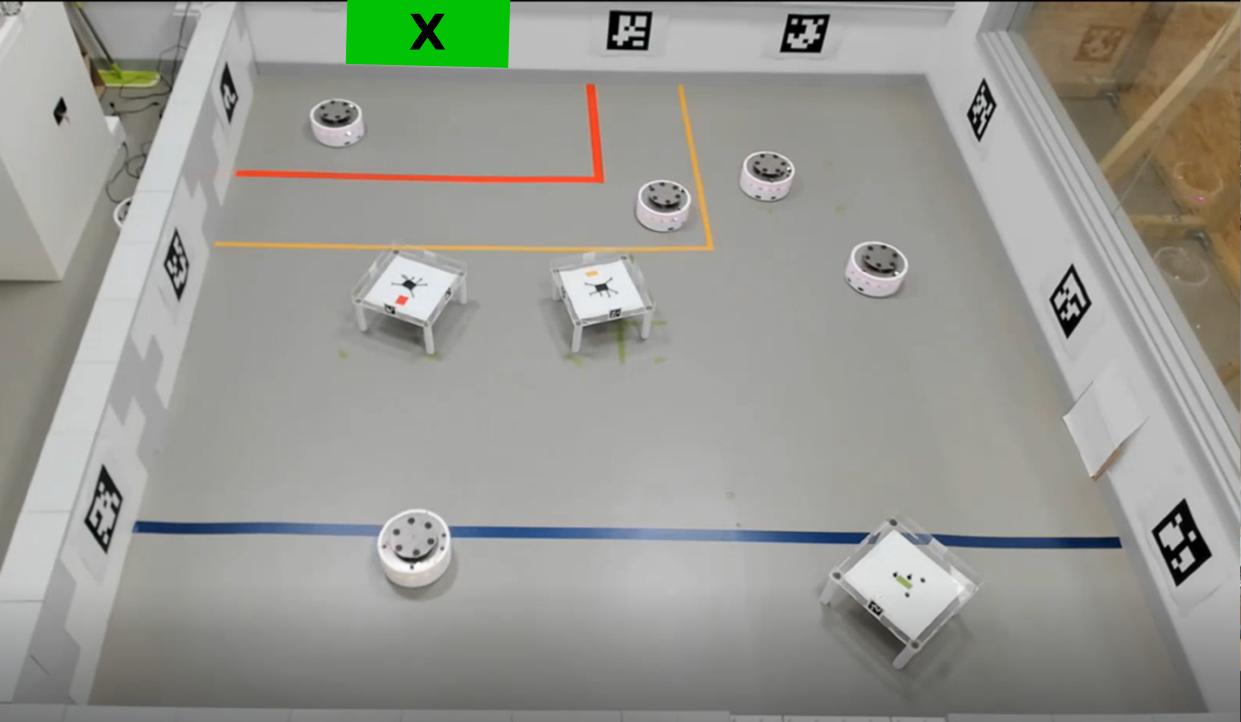}
	\caption{DOTS robots searching for carriers in the cloakroom arena. The fire exit is shaded green and marked with an X.  Tape is used to denote the blue deposit area, the red fire exit, and the amber buffer zones.} 
	\label{fig:DOTS}
\end{figure}

Here, we describe an integration test of the DOTS (Fig.~\ref{fig:DOTS}) that would form part of validation.  Unlike the previous simulations, this test uses physical hardware as in the cloakroom case study. This test mirrors the final phase of testing a developer may conduct before a system is signed off for release. While system-level tests of this kind can be used to assess confidence in multiple requirements, here we gather evidence for only the fire exit requirement.

The format of the test is as follows. The physical arena, control of the DOTS robots, and initial conditions were arranged as described in Section~\ref{sim_setup}. This ensures consistency across models (or views~\cite{iso15288}).  Due to the higher cost of conducting tests in a physical environment compared with simulation, only five trials were conducted.  Red, yellow and blue zones were marked with tape to provide a visual reference (Fig.~\ref{fig:LF_sim}), without influencing the behaviour of the robots.  The key difference from previous tests, aside from using DOTS and the lower number of trials, was that the positions of each robot were tracked using an external measurement system (OptiTrack). Each robot carried markers arranged in a unique pattern on its top surface, and the system used an array of cameras to record the position of each robot via these markers. By using an external measurement system, evidence is not biased by any potential errors in the DOTS hardware.  Measurements were streamed over a set of Robotic Operating System (ROS) topics~\cite{macenski2022} and recorded for later analysis.  

Unlike the LF and HF simulations, measurement data arrived over the ROS network at frequent but irregular intervals.  For example, equivalent samples may be recorded for one robot at $2.010$ s and $2.020$ s, and for a different robot at $2.015$ s and $2.025$ s. To ensure consistency with the previous simulations, the positional data was downsampled from circa $100 Hz$ to $1 Hz$ for each robot, using the nearest available sample to each second. This created a dataset where the location of all five robots was available at 1 s intervals. 

Analyzing the OptiTrack results of the physical experiments, we observed the following averages across the five trials: 73 s when a robot was is in the red zone, 44 s when more than one robot was in the amber zone, and 80 s when a single robot was in the amber zone. The corresponding results from the HF simulation, averaged across 10 trials, are 46 s, 11 s, and 31 s, indicating that the average time a robot spends in each zone was higher in the physical trials than in the HF simulation.

A limitation of the measurement system was that it required a robot (or more specifically, the reflectors) to be visible to multiple cameras simultaneously. Therefore, the position of a robot could be temporarily lost, for instance, when it was underneath a carrier.  This limitation is seen in the test results as longer intervals between samples. Disregarding intervals less than 1 s in length, positional data was available at worst $79\%$ and best $90\%$ of the 200 x 5 s of recorded data.  The longest interval between samples for a single robot was $22.8$ s. The raw and downsampled datasets together with the Jupyter notebooks for their processing are included in~\cite{DataRep2024}.

\section{Discussion and Conclusion}\label{sec:7}
In this section, we discuss key observations made during V\&V, identify limitations of our approach, and provide insights for future work. The LF and the HF simulations and physical tests agree that the fire exit requirement is violated. 
This consistency is crucial in development where evidence is needed to support necessary changes to a system, and for a developer, the justification to incur additional developmental costs to address them. 
When analyzing the LF and HF fire exit simulation data, we expected variations in the probabilities of a robot entering the red zone or the amber zone. These variations were anticipated due to the use of the two different simulators, and differences in the implementation of the robot controllers at the different levels of granularity (e.g., random walk, collision avoidance, localization). Our analysis confirmed these expectations. 
From the HF to the real world, as predicted, there are variations in the observed probabilities of a robot entering the red or amber zone. 
Thus, different results are expected in any single pair of trials due to the random walk behaviours of the robots and an accumulation of small inaccuracies in the simulation over a complete trial. 
We may expect the difference to reduce when the results are averaged over a large number of trials, but the relatively high cost of physical tests makes collecting the information to make this assessment infeasible. 

\textit{Verifiability} can be achieved by prioritizing verification early in the process, such as during specification and system design, where it can be elevated to a primary system design objective~\cite{Mousavi2022,Eder2021}. This approach results in systems that, by their construction, are verifiable. Based on our experiences in this work, we encountered the following opportunities to make the system more verifiable:

\noindent\textit{\textbf{Enhancing swarm behaviour to reduce the number of violations:} } Our work has identified property violations across LF and HF simulations, as well as in real-robot scenarios. The next step is to enhance swarm behaviour. For example, considering REQ 1, we could modify the environment by placing a Bluetooth beacon~\cite{Jones2022dots} indicating the fire exit door and modify the design of the robots to avoid that area. 
Regarding REQ 2, we could disperse the robots more effectively to minimize the formation of potential clusters. Subsequently, we can apply our corroborative approach to assess whether the specification is now met.

\noindent\textit{\textbf{Progressing main states over avoidance states:} } 
As mentioned in Section~\ref{sec:5.2}, during the verification of LF datasets, a key observation was that avoidance states occur more frequently than main states, which could potentially lead to livelock situations. To prevent this, we plan to introduce random timeouts to the avoidance states, aiming to prevent potential resource starvation and improve the progress of the main states in our formal models and designs.

\noindent\textit{\textbf{Explicit DROPOFF state:} } Initially, our LF simulation lacked an explicit DROPOFF state, resulting in an immediate transition from PICKUP to SEARCHING within a single iteration. In response to feedback from verification engineers, the designers introduced an explicit DROPOFF state. Moving forward, applying the lessons learned from this experience, the swarm FSM will be reviewed by the design and the verification engineers before developing the simulation.

For future work, there is significant value in utilizing the collective behaviour specified at the macroscopic level to automatically synthesize correct-by-construction, provably correct collective behaviour for the swarm. 
Such an approach would not only offer formal guarantees on fulfilling collective objectives but also aid in automating swarm design and deployment. 
However, a key challenge lies in specifying and synthesizing controllers capable of responding to dynamically changing and potentially adversarial environments~\cite{Moarref2020}. 
As an initial step towards synthesis, we will explore building a synthesizer that can take our formal properties and automatically configure or parameterize different swarm behaviours that satisfy those formal properties and the specification.

\section*{Data statement} 
All models, datasets, processing scripts and results (low-fidelity and high-fidelity simulations, formal verification, and validation) are available from the University of Bristol’s Research Data Repository~\cite{DataRep2024}.

\section*{ACKNOWLEDGMENT}

This work has been supported by the UKRI Trustworthy Autonomous Systems Node in Functionality under Grant EP/V026518/1. D.A. is currently supported by the Centre for Robotic Autonomy in Demanding and Long-lasting Environments (CRADLE) under EPSRC grant EP/X02489X/1. 


\bibliographystyle{IEEEtran}
\bibliography{SVV-bib}	

\begin{thebibliography}{10}
\providecommand{\url}[1]{#1}
\csname url@samestyle\endcsname
\providecommand{\newblock}{\relax}
\providecommand{\bibinfo}[2]{#2}
\providecommand{\BIBentrySTDinterwordspacing}{\spaceskip=0pt\relax}
\providecommand{\BIBentryALTinterwordstretchfactor}{4}
\providecommand{\BIBentryALTinterwordspacing}{\spaceskip=\fontdimen2\font plus
\BIBentryALTinterwordstretchfactor\fontdimen3\font minus
  \fontdimen4\font\relax}
\providecommand{\BIBforeignlanguage}[2]{{%
\expandafter\ifx\csname l@#1\endcsname\relax
\typeout{** WARNING: IEEEtran.bst: No hyphenation pattern has been}%
\typeout{** loaded for the language `#1'. Using the pattern for}%
\typeout{** the default language instead.}%
\else
\language=\csname l@#1\endcsname
\fi
#2}}
\providecommand{\BIBdecl}{\relax}
\BIBdecl

\bibitem{Sahin2005}
E.~{\c{S}}ahin, ``Swarm robotics: From sources of inspiration to domains of
  application,'' in \emph{Swarm Robotics}, ser. LNCS, vol. 3342.\hskip 1em plus
  0.5em minus 0.4em\relax Springer, 2005, pp. 10--20.

\bibitem{AERoS}
D.~B. Abeywickrama, J.~Wilson, S.~Lee, G.~Chance, P.~D. Winter, A.~Manzini,
  I.~Habli, S.~Windsor, S.~Hauert, and K.~Eder, ``{AERoS}: {A}ssurance of
  emergent behaviour in autonomous robotic swarms,'' in \emph{Computer Safety,
  Reliability, and Security. SAFECOMP 2023 Workshops}.\hskip 1em plus 0.5em
  minus 0.4em\relax Springer, 2023, pp. 341--354.

\bibitem{AbeyCACM2024}
D.~B. Abeywickrama, A.~Bennaceur, G.~Chance, Y.~Demiris, A.~Kordoni, M.~Levine,
  L.~Moffat, L.~Moreau, M.~R. Mousavi, B.~Nuseibeh, S.~Ramamoorthy, J.~O.
  Ringert, J.~Wilson, S.~Windsor, and K.~Eder, ``On specifying for
  trustworthiness,'' \emph{Commun. ACM}, vol.~67, no.~1, p. 98–109, dec 2023.

\bibitem{Webster2020}
M.~Webster, D.~Western, D.~Araiza-Illan, D.~Clare, K.~Eder, M.~Fisher, and
  A.~Pipe, ``A corroborative approach to verification and validation of
  human–robot teams,'' \emph{The International Journal of Robotics Research},
  vol.~39, no.~1, pp. 73--99, 2020.

\bibitem{Jones2022dots}
S.~Jones, E.~Milner, M.~Sooriyabandara, and S.~Hauert, ``\relax{DOTS: An open
  testbed for industrial swarm robotic solutions},'' \emph{arXiv}, 2022.

\bibitem{Liu2010}
W.~Liu and A.~F.~T. Winfield, ``Modeling and optimization of adaptive foraging
  in swarm robotic systems,'' \emph{The International Journal of Robotics
  Research}, vol.~29, no.~14, pp. 1743--1760, 2010.

\bibitem{Konur2012}
S.~Konur, C.~Dixon, and M.~Fisher, ``Analysing robot swarm behaviour via
  probabilistic model checking,'' \emph{Robot. Auton. Syst.}, vol.~60, no.~2,
  p. 199–213, feb 2012.

\bibitem{Lerman2005}
K.~Lerman, A.~Martinoli, and A.~Galstyan, ``A review of probabilistic
  macroscopic models for swarm robotic systems,'' in \emph{Swarm Robotics},
  E.~{\c{S}}ahin and W.~M. Spears, Eds.\hskip 1em plus 0.5em minus 0.4em\relax
  Springer, 2005, pp. 143--152.

\bibitem{Winfield2005}
A.~F. Winfield, J.~Sa, M.-C. Fernández-Gago, C.~Dixon, and M.~Fisher, ``On
  formal specification of emergent behaviours in swarm robotic systems,''
  \emph{International Journal of Advanced Robotic Systems}, vol.~2, no.~4,
  p.~39, 2005.

\bibitem{Dixon2011}
C.~Dixon, A.~Winfield, and M.~Fisher, ``Towards temporal verification of
  emergent behaviours in swarm robotic systems,'' in \emph{Towards Autonomous
  Robotic Systems}.\hskip 1em plus 0.5em minus 0.4em\relax Springer, 2011, pp.
  336--347.

\bibitem{Endo2023}
W.~Endo, C.~Baumann, H.~Asama, and A.~Martinoli, ``Automatic multi-robot
  control design and optimization leveraging multi-level modeling: An
  exploration case study,'' \emph{IFAC-PapersOnLine}, vol.~56, no.~2, pp.
  11\,462--11\,469, 2023.

\bibitem{Egerstedt1999}
M.~Egerstedt, K.~Johansson, J.~Lygeros, and S.~Sastry, ``Behavior based
  robotics using regularized hybrid automata,'' in \emph{Proc. of the 38th IEEE
  Conference on Decision and Control}, vol.~4, 1999, pp. 3400--3405.

\bibitem{Martinoli2004}
A.~Martinoli, K.~Easton, and W.~Agassounon, ``Modeling swarm robotic systems: a
  case study in collaborative distributed manipulation,'' \emph{The
  International Journal of Robotics Research}, vol.~23, no. 4-5, pp. 415--436,
  2004.

\bibitem{Brambilla2013}
M.~Brambilla, E.~Ferrante, M.~Birattari, and M.~Dorigo, ``Swarm robotics: a
  review from the swarm engineering perspective,'' \emph{Swarm Intelligence},
  vol.~7, no.~1, pp. 1--41, 2013.

\bibitem{Hawkins2021}
R.~Hawkins, C.~Paterson, C.~Picardi, Y.~Jia, R.~Calinescu, and I.~Habli,
  ``\relax{Guidance on the assurance of machine learning in autonomous systems
  (AMLAS)},'' University of York, Guidance Version 1.1, Mar. 2021.

\bibitem{Baumann2022}
C.~Baumann, H.~Birch, and A.~Martinoli, ``Leveraging multi-level modelling to
  automatically design behavioral arbitrators in robotic controllers,'' in
  \emph{Proc. of the 2022 IEEE/RSJ International Conference on Intelligent
  Robots and Systems (IROS)}, 2022, pp. 9318--9325.

\bibitem{Lee2022}
S.~Lee, E.~Milner, and S.~Hauert, ``A data-driven method for metric extraction
  to detect faults in robot swarms,'' \emph{IEEE Robot. Autom. Lett.}, vol.~7,
  no.~4, pp. 10\,746--10\,753, 2022.

\bibitem{Lee2023}
S.~Lee and S.~Hauert, ``Building trustworthiness by minimizing the sim-to-real
  gap in fault detection for robot swarms,'' in \emph{Proc. of the First
  International Symposium on Trustworthy Autonomous Systems (TAS'23)}.\hskip
  1em plus 0.5em minus 0.4em\relax ACM, 2023.

\bibitem{PRISM_P}
\BIBentryALTinterwordspacing
\relax{PRISM Model Checker}, ``\relax{Property specification},'' Online, 2023.
  [Online]. Available:
  \url{https://www.prismmodelchecker.org/manual/PropertySpecification/AllOnOnePage}
\BIBentrySTDinterwordspacing

\bibitem{Jones2020}
S.~Jones, E.~Milner, M.~Sooriyabandara, and S.~Hauert, ``Distributed
  situational awareness in robot swarms,'' \emph{Advanced Intelligent Systems},
  vol.~2, no.~11, p. 2000110, 2020.

\bibitem{Dougherty1995}
J.~Dougherty, R.~Kohavi, and M.~Sahami, ``Supervised and unsupervised
  discretization of continuous features,'' in \emph{Machine Learning
  Proceedings 1995}.\hskip 1em plus 0.5em minus 0.4em\relax Morgan Kaufmann,
  1995, pp. 194--202.

\bibitem{ieee2012}
{The Institute of Electrical and Electronics Engineers (IEEE)}, ``{IEEE}
  standard for system, software, and hardware verification and validation,''
  \emph{IEEE 1012-2016}, pp. 1--260, 2017.

\bibitem{iso15288}
{International Organization for Standardization (ISO)}, ``Systems and software
  engineering--system life cycle processes,'' \emph{ISO/IEC/IEEE
  15288:2023(E)}, pp. 1--128, 2023.

\bibitem{macenski2022}
S.~Macenski, T.~Foote, B.~Gerkey, C.~Lalancette, and W.~Woodall, ``Robot
  operating system 2: Design, architecture, and uses in the wild,''
  \emph{Science Robotics}, vol.~7, no.~66, p. eabm6074, 2022.

\bibitem{DataRep2024}
D.~B. Abeywickrama, S.~Lee, C.~Bennett, R.~Abu-Aisheh, T.~Didiot-Cook,
  S.~Jones, S.~Hauert, and K.~Eder, ``Autonomous robotic swarms: A
  corroborative approach for verification and validation: Models, datasets and
  results,'' \url{https://doi.org/10.5523/bris.16ckwq1v44odw29sbsvca8yyhs}.

\bibitem{Mousavi2022}
M.~R. Mousavi, A.~Cavalcanti, M.~Fisher, L.~Dennis, R.~Hierons, B.~Kaddouh,
  E.~L.-C. Law, R.~Richardson, J.~O. Ringer, I.~Tyukin, and J.~Woodcock,
  ``Trustworthy autonomous systems through verifiability,'' \emph{Computer},
  vol.~56, no.~2, pp. 40--47, 2023.

\bibitem{Eder2021}
K.~Eder, \emph{Gaining Confidence in the Trustworthiness of Robotic and
  Autonomous Systems}.\hskip 1em plus 0.5em minus 0.4em\relax Cham: Springer,
  2021, pp. 139--164.

\bibitem{Moarref2020}
S.~Moarref and H.~Kress-Gazit, ``Automated synthesis of decentralized
  controllers for robot swarms from high-level temporal logic specifications,''
  \emph{Autonomous Robots}, vol.~44, no. 3-4, pp. 580--600, 2020.

\end{thebibliography}

\end{document}